# A software for aging faces applied to ancient marble busts

**Amelia Carolina Sparavigna**
Department of Applied Science and Technology, Politecnico di Torino, Italy

The study and development of software able to show the effect of aging of faces is one of the tasks of face recognition technologies. Some software solutions are used for investigations, some others to show the effects of drugs on healthy appearance, however some other applications can be proposed for the analysis of visual arts. Here we use a freely available software, which is providing interesting results, for the comparison of ancient marble busts. An analysis of Augustus busts is proposed.

Face recognition software and technologies are focusing on the possibility of computer algorithms to recognize a face in some galleries of images, which can be acquired from pictures provided by still images or frames of a video sequence. The software algorithms then must reproduce the innate human ability to recognize a face, a fundamental task in mimic the behavior of the human brain. Another important research in the field of face recognition is the study of the effects of aging in the craniofacial morphology [1]. Human faces change during the life, their features varying affected by several factors ranging from the inherent genetics and the environmental constrains.

As told in Ref.2, several agencies of investigation regularly require matching a probe image with the individuals in the missing person database. However, there are often significant differences between facial features of probe and gallery images due to age variation. For instance if the probe image is a 15 years-old boy or girl and the gallery image of the same person is of 5 years, the face recognition algorithm must perform a very difficult task. Researchers have then proposed several age simulation and modeling techniques [2]. These models alter the face according to the facial growth over a specific period of time. The reader can find several references given in [2], the oldest is that of Burt and Perrett [3]. Since the face aging is affecting the performance of face recognition systems, the analysis of synthetically generating age-progressed or age-regressed images is a good method of improving the robustness of face-based biometrics [1]. In Ref.4, the accuracy of methods for the security of biometric verification systems is investigated. The paper presents methods of modeling and predicting facial template aging based on matching score analysis.

An interesting social application of a software solution for aging faces was proposed by the Task Force for Tobacco-Free Women and Girls in New York State, which utilized it to illustrate how smoking can affect the facial appearance [5]. The task force members reviewed the literature on the association between smoking and facial wrinkling, provided parameters for customization of the APRIL (age progression image launcher) [6]. Photoshop is also used for ageing the faces, using its FaceAge® plugin. However, this is not freely available.

Some software solutions are then used for investigation, some others to show the effects of drugs on healthy appearance, however other applications can be imagined and used, for instance, in the analysis of visual arts. A web site [7], in20years.com, allows to have a freely available processing of images, to test the possibility to use it for these purposes. In20years displays how a person looks after several years. According to the website, "In20years is using advanced face detection and morphing technology to predict what your face would look like in 20 or 30 years from now." As options, it is possible to select gender and the use of drugs or not. Afterwards, one uploads a picture and after a few seconds the site gives an image of an older face. It is a site for fun [8], however, using a picture which is showing a full face, it provides interesting results, in particular, as we can see, on the pictures of marble busts.

To analyze busts, I have recently suggest [9] the use of a 3D rendering to create some virtual copies of ancient artifacts to study and compare them. In particular, this approach could be interesting for the portraitures of Julius Caesar. In general, the studies of ancient busts is important to identify the person portrayed in the marble and the epoch the artist created it.

Let me show here an interesting analysis where the aging software can be applied: this is the comparison of the Roman emperor Augustus busts, portrayed when he was young and his busts as a man. In the Figure 1 on the right, we can see two busts showing Augustus as a child. On the left, the images obtained after processing on the In20years site, choosing the appearance of the boy after twenty years. The result is quite good and realistic. However, we could ask ourselves whether this is correct or not. Let us compare the processing with the faces of two statues, one shown in Figure 2 at Louvre, Paris, and that in the Figure 3, the Via Labicana Augustus, dated 12 BC. The lower face in the Figure 1 looks like that in Figure 2, and the upper one that of Figure 3.

As a conclusion, we can tell that the freely available software for aging faces can be easily applied to marble busts to obtain a more realistic image of the portrayed person. Moreover, it can be useful to compare the portraitures of the same person at different ages, as shown by the analysis of Augustus busts, or to solve some problems concerning her/his actual identification.

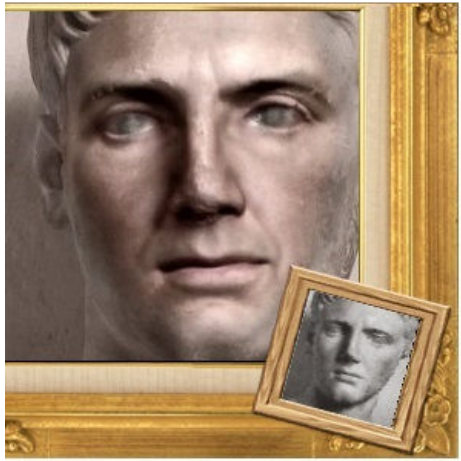
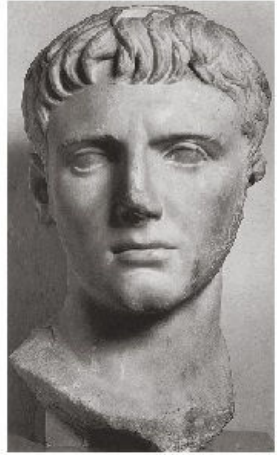
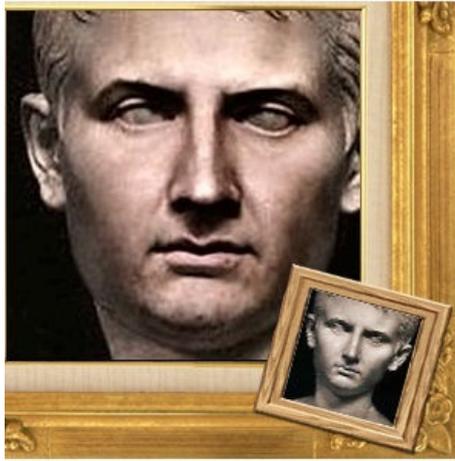
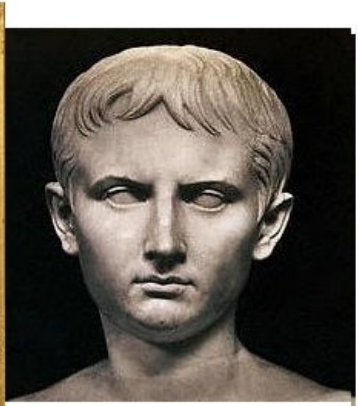

Figure 1: On the right the busts of a young Augustus and on the left the images obtained after processing them using the freely available software of In20Years.com. The two resulting images are rather interesting when compared with the following images of Augusts as a man.

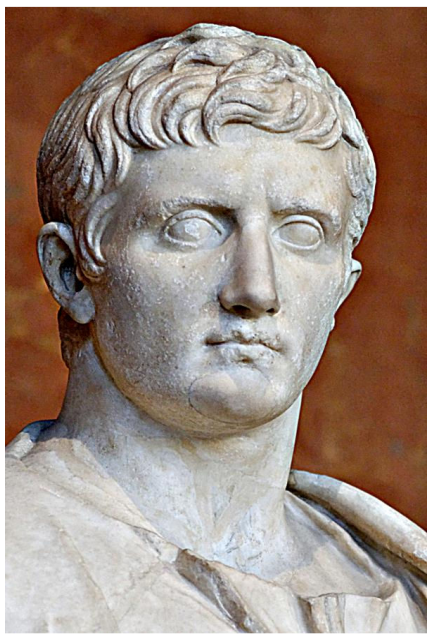
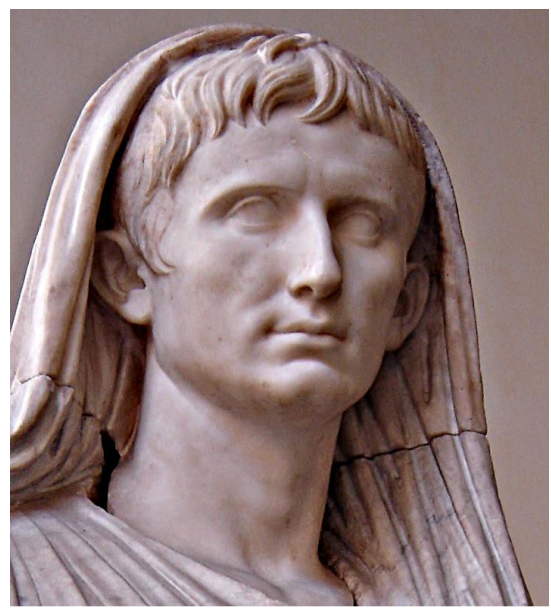

Fig.2 Augustus at Louvre, Paris

Fig.3 Augustus of Via Labicana.